\documentclass{article}

\usepackage{microtype}
\usepackage{graphicx}
\usepackage{subcaption}
\usepackage{booktabs}
\usepackage{hyperref}

\usepackage[accepted]{icml2026}

\usepackage{amsmath}
\usepackage{amssymb}
\usepackage{amsfonts}
\usepackage{amsthm}
\usepackage{bm}

\newtheorem{theorem}{Theorem}

\usepackage{xcolor}
\usepackage{multicol}
\usepackage{multirow}
\usepackage{wrapfig}
\usepackage{adjustbox}
\usepackage{enumitem}
\usepackage{array}
\usepackage{pifont}
\usepackage{balance}

\newcommand{\se}[2]{$#1_{\pm #2}$}
\newcommand{\seb}[2]{$\mathbf{#1}_{\pm #2}$}

\makeatletter
\g@addto@macro\UrlBreaks{\do\/\do\.\do\-\do\_\do\:\do\?\do\=\do\&\do\#\do\%%
\do\a\do\b\do\c\do\d\do\e\do\f\do\g\do\h\do\i\do\j\do\k\do\l\do\m\do\n\do\o\do\p\do\q\do\r\do\s\do\t\do\u\do\v\do\w\do\x\do\y\do\z%
\do\A\do\B\do\C\do\D\do\E\do\F\do\G\do\H\do\I\do\J\do\K\do\L\do\M\do\N\do\O\do\P\do\Q\do\R\do\S\do\T\do\U\do\V\do\W\do\X\do\Y\do\Z%
\do\0\do\1\do\2\do\3\do\4\do\5\do\6\do\7\do\8\do\9}
\makeatother

\icmltitlerunning{ReTabSyn: Realistic Tabular Data Synthesis via RL}

\begin{document}

\twocolumn[
\icmltitle{ReTabSyn: Realistic Tabular Data Synthesis via Reinforcement Learning}

\begin{icmlauthorlist}
\icmlauthor{Xiaofeng Lin}{ucla}
\icmlauthor{Seungbae Kim}{usf}
\icmlauthor{Zhuoya Li}{ucla}
\icmlauthor{Zachary DeSoto}{ucla}
\icmlauthor{Charles Fleming}{cisco,um}
\icmlauthor{Guang Cheng}{ucla}
\end{icmlauthorlist}

\icmlaffiliation{ucla}{University of California, Los Angeles, USA}
\icmlaffiliation{usf}{University of South Florida, USA}
\icmlaffiliation{cisco}{Cisco, USA}
\icmlaffiliation{um}{University of Mississippi, USA}

\icmlcorrespondingauthor{Guang Cheng}{guangcheng@stat.ucla.edu}

\icmlkeywords{Tabular Data, Synthetic Data, Reinforcement Learning}

\vskip 0.3in
]

\printAffiliationsAndNotice{}

\begin{abstract}
Deep generative models can help with data scarcity and privacy by producing synthetic training data, but they struggle in low-data, imbalanced tabular settings to fully learn the complex data distribution. 
We argue that striving for the full joint distribution could be overkill; for greater data efficiency, models should prioritize learning the conditional distribution $P(y\mid \bm{X})$, as suggested by recent theoretical analysis.
Therefore, we overcome this limitation with \textbf{ReTabSyn}, a \textbf{Re}inforced \textbf{Tab}ular \textbf{Syn}thesis pipeline that provides direct feedback on feature correlation preservation during synthesizer training. 
This objective encourages the generator to prioritize the most useful predictive signals when training data is limited, thereby strengthening downstream model utility. 
We empirically fine-tune a language model-based generator using this approach, and across benchmarks with small sample sizes, class imbalance, and distribution shift, ReTabSyn consistently outperforms state-of-the-art baselines. 
Moreover, our approach can be readily extended to control various aspects of synthetic tabular data, such as applying expert-specified constraints on generated observations.
\end{abstract}


\section{Introduction}

\begin{figure}[th]
    \centering    
    \includegraphics[width=\linewidth]{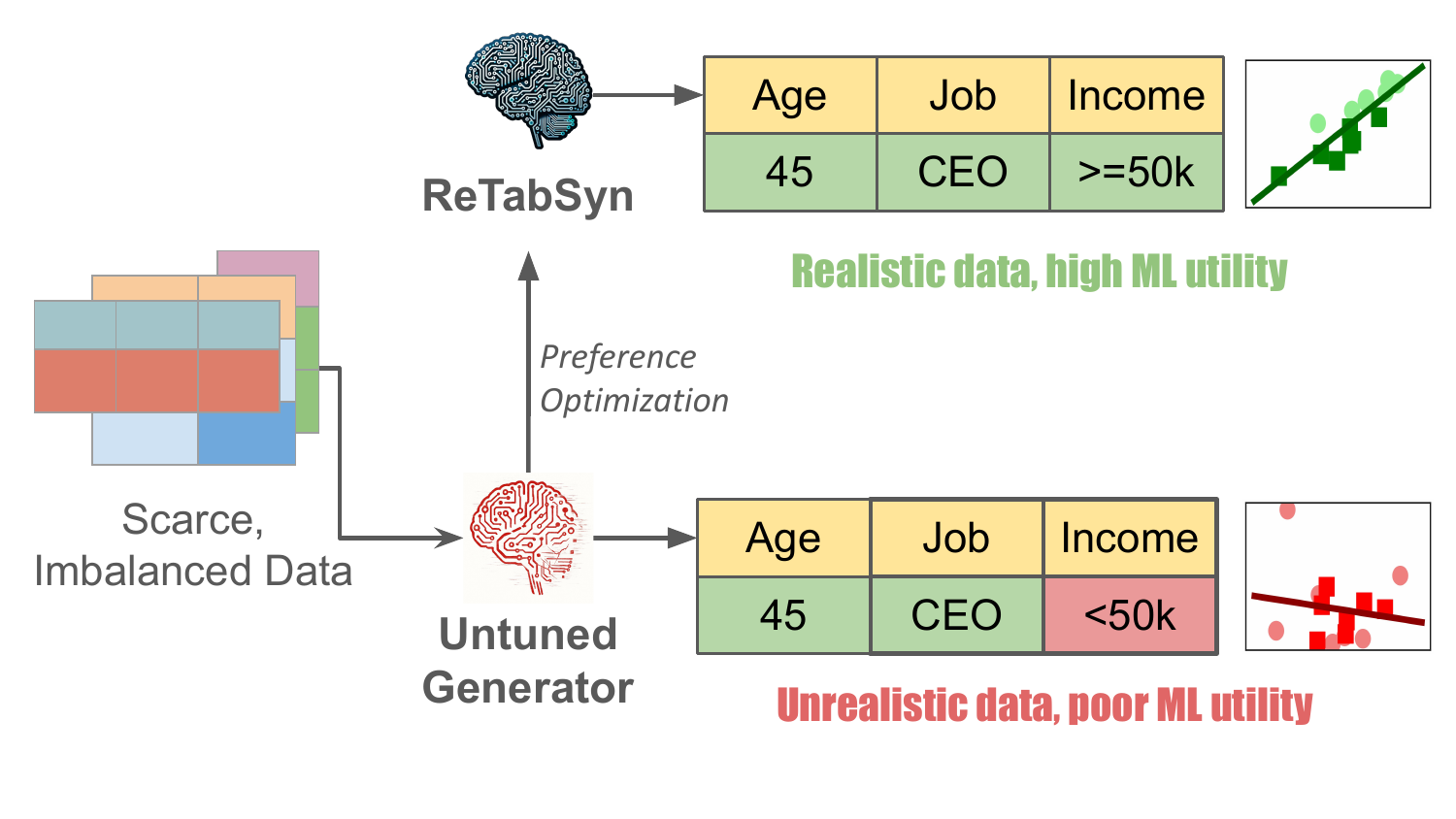}
    \caption{Illustration of the challenging scenario for tabular generator: in scenarios with limited or imbalanced training data, tabular generators may produce synthetic datasets containing unrealistic entries, potentially degrading the performance of downstream machine learning tasks.
    }
    \label{fig:overview}
\end{figure}

Deep Generative Models (DGMs) have recently gained significant attention for their ability to generate high-quality data for model training in various domains, particularly in text and vision~\cite{rombach2022stablediffusion,li2024pretextgen,blattmann2023align}. Synthetic data generated by DGMs has been explored as a potential solution to data scarcity and privacy constraints, aiming to augment machine learning (ML) models and improve generalization across tasks~\cite{whitehouse-etal-2023-llm,dai2023auggpt, trabuccoeffective}. Among the various data modalities, tabular data synthesis is especially critical because structured datasets underpin decision-making in essential sectors such as healthcare, finance, and social science, where data is often limited by strict privacy constraints and regulatory restrictions. Various tabular DGMs, including Generative Adversarial Networks (GANs), Variational Autoencoders (VAEs)~\cite{xu2019modeling,zhao2022ctabplus}, diffusion models~\cite{kotelnikov2023tabddpm,zhang2023tabsyn}, and autoregressive language models~\cite{borisov2022language}, have been proposed and shown strong performance.

Despite these advancements, in practical scenarios involving small, imbalanced, or distribution-shifted datasets, tabular DGMs often produce synthetic data that lacks the desired high downstream utility and augmentation usefulness. Figure~\ref{fig:overview} illustrates a case where the generated data deviates from expected norms:for example, a model trained on scarce data may erroneously generate a record showing a CEO earning less than \$50K.



We argue that this failure stems from a fundamental misalignment between the generative objective and the requirements for downstream utility. Standard DGMs attempt to learn the full joint distribution $P(\bm{X}, y)$, a data-intensive task that becomes intractable in sparse regimes. However, recent theoretical advancements in synthetic data utility suggest that perfect joint distribution matching is unnecessary. Specifically, \cite{xu2023utility} proved that a synthesizer can achieve \emph{zero loss} of downstream performance relative to real data provided it accurately models the conditional distribution $P(y \mid \bm{X})$, even if the marginal feature distribution is imperfect. This theoretical insight suggests that to maximize utility in low-data settings, the generative process must prioritize the preservation of decision boundaries over generic distributional fidelity.

Guided by this theoretical principle, we introduce \textbf{ReTabSyn} (\textbf{Re}inforced \textbf{Tab}ular \textbf{Syn}thesis), a framework designed to explicitly enforce the priority of $P(y \mid \bm{X})$. Rather than relying solely on likelihood maximization which treats all correlations equally, we align a pre-trained tabular generator to decision-relevant structures using \emph{Direct Preference Optimization} (DPO).
We construct \emph{chosen–rejected preference pairs} by applying perturbations to real rows to create direct feedback signals:  
(i) \emph{Target Perturbation:} modifying $y$ supplies a margin signal that sharpens the desired conditional $P_\theta(y\mid \bm{X})$, directly addressing the theoretical requirement for utility;  
(ii) \emph{Predictor Perturbation:} modifying predictor features $X$ penalizes unrealistic feature–feature co-occurrences to maintain sufficient structural consistency.
By enlarging the log-likelihood margin $\log \pi_\theta(y\mid \bm{x})-\log \pi_\theta(\tilde{y}\mid \bm{x})$ on these pairs, ReTabSyn effectively "locks in" the conditional relationships essential for downstream tasks, even when data volume is insufficient to learn the full joint distribution.

Unlike classic \emph{reinforcement learning with human feedback} (RLHF) for LLMs \cite{ouyang2022instructGPT,NEURIPS2023_a85b405e} and recent tabular RL approaches that rely on an external \emph{oracle} classifier~\cite{yang2024pta, das2024synrl}, \textbf{ReTabSyn is both oracle-free and human-label-free}.  
Structured tables permit reliable, rule-based perturbations (type checks, monotonic constraints, logical rules) that yield \emph{high-purity supervision} at scale.
This tabular-native advantage removes guidance-model bias, lowers privacy and compute cost, and, as we show empirically, closes the utility gap between synthetic and real data in low-data, imbalanced, and distribution-shift settings.

We summarize our core contributions as follows:
\begin{itemize}[leftmargin=3mm]
    \item \textbf{Oracle-free, tabular-native preference construction.} 
    We introduce a target-consistent perturbation strategy that yields chosen–rejected pairs for DPO, eliminating any external reward/oracle model or human labels. We release all code, runtime and configs for full reproducibility.\footnote{\url{https://github.com/Bernardo1998/ReTabSyn}}

    \item \textbf{Decision-focused conditional alignment.} 
    We fine-tune a pre-trained tabular generator with DPO on these pairs. This approach is theoretically motivated to prioritize the conditional $P_\theta(y\mid \bm{X})$ while preserving key feature–feature dependencies, thereby closing the utility gap in low-data and rare-event regimes. 

    \item \textbf{Robust benchmarks and diagnostics.} 
    We evaluate on challenging settings (e.g., $0.5\%$ positive-rate tasks, 10 random seeds) and introduce comprehensive downstream utility, marginal distribution, and joint-distribution fidelity metrics to certify the advantage of our method under realistic scenarios of small, imbalanced, and distribution-shifted data.

\end{itemize}

\paragraph{Conflict of Interest Disclosure.}
This research was supported in part by a gift from Cisco, the National Science Foundation (NSF CNS-2247795), and the Office of Naval Research (ONR N00014-22-1-2680). One of the authors is employed by Cisco. We declare no conflict of interest: the models and methods evaluated in this paper are not developed or commercialized by any of these funding parties, so our evaluation does not assess any funder's product.

\section{Related Work}

\subsection{Synthetic Tabular Data Generation}
Tabular synthesis has progressed from classical resampling/model-based approaches~\cite{chawla2002smote, nowok2016synthpop} to deep generative models, including GAN/VAE methods (e.g., CTGAN/TVAE/CtabGAN+)~\cite{xu2019modeling, zhao2022ctabplus}, diffusion-based models (TabDDPM, TabSyn)~\cite{kotelnikov2023tabddpm, zhang2023tabsyn}, and autoregressive language-model generators (GReaT, CuratedLLM)~\cite{borisov2022language, seedat2023curated}. Privacy-oriented mechanisms (e.g., PATE-GAN, AIM) offer formal guarantees but often reduce downstream utility~\cite{jordon2018pate, mckenna2022aim}. Recent work also studies controllability and constraint satisfaction for tabular synthesis~\cite{vero2024cuts, howRealistic}.

Most prior methods optimize a density objective for the joint distribution (or $P_\theta(\bm{X}\mid y)$), which can spend limited statistical budget on marginal realism instead of decision-relevant structure in small/imbalanced regimes. ReTabSyn differs by aligning the generator to the conditional $P_\theta(y\mid\bm{X})$: we construct oracle-free, high-purity preference pairs via schema-validated perturbations of real rows and fine-tune with DPO~\cite{rafailov2023direct}, while also discouraging unrealistic feature–feature co-occurrences.




\subsection{Reinforcement Learning for Deep Generative Models}
Reinforcement learning and preference optimization are widely used to align generative models to desired properties. In RLHF, a reward model trained from preference rankings is optimized with Proximal Policy Optimization (PPO)~\cite{schulman2017proximal, NEURIPS2022_b1efde53}, and RLAIF reduces human annotation cost by using AI-generated preferences~\cite{bai2022constitutional, lee2023rlaif}. DPO provides a reward-model-free alternative that learns directly from preference pairs~\cite{rafailov2023direct}.

In tabular synthesis, PTA and SynRL apply RL or preference optimization to improve utility or enforce constraints, but typically rely on external guidance models (e.g., task classifiers or reward models) to score samples~\cite{yang2024pta, das2024synrl}. In contrast, ReTabSyn is oracle-free and human-label-free: we exploit tabular schemas to generate target-consistent perturbations that yield accepted–rejected pairs at scale, avoiding guidance-model bias from imperfect oracles.

\begin{figure*}[t]
    \centering
    \includegraphics[width=\textwidth]{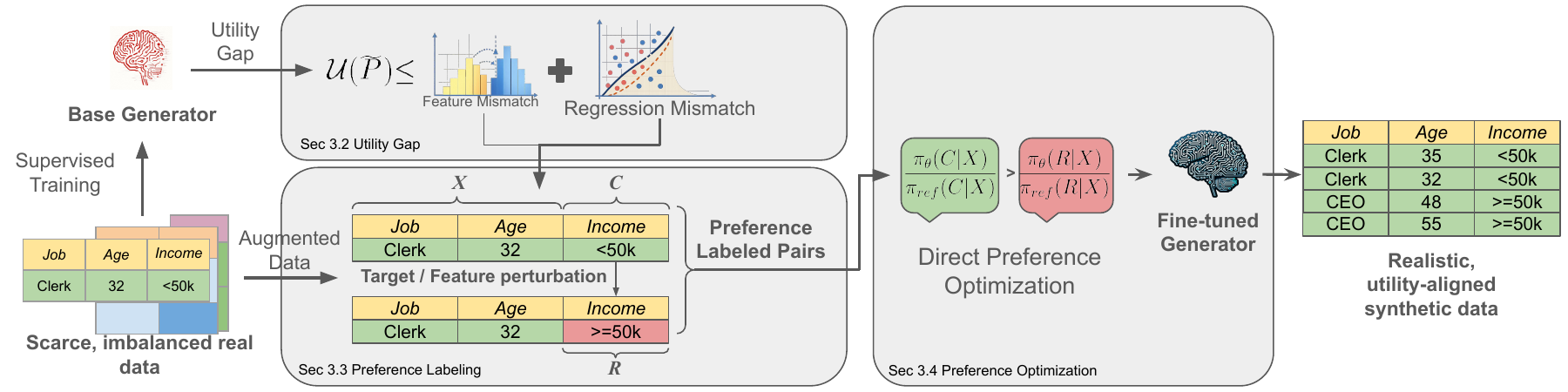}
    \caption{
        Overall workflow of ReTabSyn. Starting from scarce, imbalanced real data, we first train a base tabular generator $\pi_{\mathrm{ref}}$ via supervised learning.
        Guided by the utility decomposition (Thm.~3.1), we construct oracle-free preference-labeled tuples by \textbf{label/target perturbation}:
        for each row, we keep the conditioning context (prompt) fixed and create a \emph{chosen} tuple with the original label and a \emph{rejected} tuple with a perturbed label, forming prompt--chosen--rejected training pairs.
        Finally, we fine-tune the generator with \textbf{DPO}, using $\pi_{\mathrm{ref}}$ as the reference policy, to enlarge the likelihood margin between chosen and rejected tuples, improving feature--target alignment and downstream ML utility.
    }
    \label{fig:rltf}
\end{figure*}

\section{Methodology}

Figure~\ref{fig:rltf} illustrates the ReTabSyn framework. We begin by formalizing the problem of tabular synthesis and providing a theoretical analysis of the utility gap, demonstrating why prioritizing conditional alignment maximizes downstream performance (Sections~\ref{sec:problem} and \ref{sec:theory}). Guided by this theory, we detail our strategy for constructing preference pairs by perturbing target and feature values to penalize regression and feature mismatches (Section~\ref{sec:preference}). We then explain how we fine-tune the model using DPO to enforce these structural constraints (Section~\ref{sec:dpo}). Finally, we describe the base generator architecture and the data augmentation technique used to initialize the training process (Section~\ref{sec:preliminaries}).

\subsection{Problem Statement}
\label{sec:problem}

Let $\mathcal{D}=\{(\bm{x}_i, y_i)\}_{i=1}^N$ be a dataset sampled i.i.d. from a true distribution $P(\bm{X}, y)$, where $\bm{X} \in \mathcal{X} \subseteq \mathbb{R}^d$ are features and $y \in \mathcal{Y}$ is the target. We use $\bm{X}$ to denote the feature vector and $y$ the target label; lowercase $\bm{x}$ denotes a feature realization. We focus on classification, though the method inherently extends to regression.
Let $\mathcal{F}$ be a hypothesis class of downstream classifiers (e.g., Random Forests, MLPs). Ideally, we seek a generator $G_\theta$ producing synthetic data $\tilde{\mathcal{D}} \sim \tilde{P}_\theta(\bm{X}, y)$ such that a classifier $\tilde{f} \in \mathcal{F}$ trained on $\tilde{\mathcal{D}}$ minimizes the risk on real data, $R(\tilde{f}) = \mathbb{E}_{P}[\ell(\tilde{f}(\bm{X}), y)]$.

Standard generative models maximize the joint log-likelihood $\mathbb{E}_{\mathcal{D}}[\log \tilde{P}_\theta(\bm{x}, y)]$. However, this objective treats the modeling of marginal features $P(\bm{X})$ and conditional labels $P(y\mid\bm{X})$ with equal importance. In regimes of data scarcity (small $N$), attempting to learn the full structure of $P(\bm{X})$ consumes the limited statistical budget, often at the expense of the decision boundary $P(y\mid\bm{X})$. We argue that to maximize downstream utility, the generation process must prioritize the conditional distribution.

\subsection{Minimizing Utility Gap}
\label{sec:theory}

We motivate our method design by analyzing the \emph{population utility gap}, defined as the excess risk incurred by training on synthetic data versus real data:
\begin{equation}
    \mathcal{U}(\tilde{P}) := R(\tilde{f}^*) - R(f^*),
\end{equation}
where $f^*$ and $\tilde{f}^*$ are the population-optimal classifiers for $P$ and $\tilde{P}$ respectively. To understand how to minimize this gap, we leverage the utility bound derived by \cite{xu2023utility}. Let $\eta(\bm{x}) = P(y=1|\bm{x})$ and $\tilde{\eta}(\bm{x}) = \tilde{P}(y=1|\bm{x})$ be the real and synthetic regression functions (conditional distributions).

\begin{theorem}[Utility Decomposition \cite{xu2023utility}]
\label{thm:utility}
Assuming loss $\ell$ is Lipschitz and bounded, the utility gap is bounded by two distinct error terms:
\begin{equation}
\label{eq:utility_bound}
    \mathcal{U}(\tilde{P}) \;\le\; \underbrace{C_1 \cdot d_{\mathcal{F}}(P_{\bm{X}}, \tilde{P}_{\bm{X}})}_{\text{Feature Mismatch}} \;+\; \underbrace{C_2 \cdot \|\tilde{\eta} - \eta\|_{L^2(P_{\bm{X}})}}_{\text{Regression Mismatch}},
\end{equation}
where $d_{\mathcal{F}}$ is an integral probability metric measuring the distance between feature marginals, and constants $C_1, C_2$ depend on the hypothesis class $\mathcal{F}$.
\end{theorem}

We stress that Theorem~\ref{thm:utility} is a prior result of \cite{xu2023utility} and not a contribution of this work; our contribution is to translate this bound into concrete training signals (Sections~\ref{sec:preference}--\ref{sec:dpo}), targeting the regression-mismatch term with Type~I perturbations and the feature-mismatch term with Type~II perturbations. Equation~\ref{eq:utility_bound} provides the theoretical grounding for \textbf{ReTabSyn} and answers two critical design questions:

\paragraph{Why prioritize rewarding $P(y\mid\bm{X})$?}
The Target Mismatch term $\|\tilde{\eta} - \eta\|$ acts as the primary factor in the utility bound. While the Feature Mismatch term $d_{\mathcal{F}}(P_{\bm{X}}, \tilde{P}_{\bm{X}})$ contributes to the gap, its impact is contingent on the downstream model class $\mathcal{F}$: as shown in \cite{xu2023utility}, if $\mathcal{F}$ is well-specified (sufficiently expressive), the constant $C_1$ becomes negligible. Prioritizing the conditional distribution is thus the preferred strategy. A failure to model the conditional $\eta$ (the ``ground truth'' logic) leads to almost certain downstream failure for \emph{all} hypothesis classes, whereas feature mismatch is only detrimental to rigid, misspecified models. Thus, minimizing $\|\tilde{\eta} - \eta\|$ represents a robust, minimax strategy that maximizes potential utility across unknown downstream tasks. Our preference optimization specifically targets this term by rewarding synthetic samples that adhere to the true conditional logic.

\paragraph{Why is this crucial for small data?}
Estimating a high-dimensional, mixed-type joint distribution is data-hungry and high-variance in small-$N$ regimes, whereas downstream predictive utility for a fixed target can be improved by directly optimizing decision-relevant conditional structure, consistent with Vapnik's principle \cite{vapnik2013nature}. In our LM-based generator, the joint likelihood decomposes into many per-column (token) prediction losses; with limited data, optimizing all columns uniformly can under-emphasize boundary-critical signal. By constructing preference pairs that hold $\bm{X}$ fixed and contrast correct vs.\ perturbed $y$, and optimizing them with DPO (Section~\ref{sec:dpo}), we inject a targeted training signal that prioritizes conditional alignment; we validate this budget trade-off empirically in Section~\ref{sec:ablation} (Figure~\ref{fig:cond_priority}).

\subsection{Preference Labeling for Utility Maximization}
\label{sec:preference}

To bridge the gap between the generative objective and downstream utility, we construct preference pairs $(C, R)$—where $C$ is a \emph{chosen} (real/augmented) row and $R$ is a \emph{rejected} (perturbed) row. Guided by Theorem~\ref{thm:utility}, we design two specific types of perturbations to minimize the theoretical error terms.

\paragraph{Type I: Target Perturbation (Minimizing Target Mismatch).}
Since our theoretical analysis identifies the conditional distribution $P(y\mid\bm{X})$ as the priority for utility, our primary supervision signal targets the relationship between features and the class label.
Formally, given a real row $(\bm{X}, y)$, we keep the chosen row $C=(\bm{X}, y)$ and form the rejected row by resampling the target,
\begin{equation}
\label{eq:typeI}
\tilde{y}\sim P(y),\quad \tilde{y}\neq y,\qquad R=(\bm{X}, \tilde{y}),
\end{equation}
where $P(y)$ is the empirical marginal over labels.
By preferring $C$ over $R$, we directly penalize the generator for assigning incorrect labels given a set of features, thereby minimizing the regression mismatch term $\|\tilde{\eta} - \eta\|$ in Eq.~\ref{eq:utility_bound}.

\paragraph{Type II: Feature Perturbation (Minimizing Feature Mismatch).}
While secondary, the feature distribution mismatch $d_{\mathcal{F}}(P_{\bm{X}}, \tilde{P}_{\bm{X}})$ also contributes to the utility gap. To maintain realistic feature structures learned during pre-training, we identify strongly correlated feature pairs $(A, B)$ using Pearson’s correlation (numeric) or Cramér’s $V$ (categorical). Formally, for such a pair we resample one member from its marginal (e.g., shifting a numeric value to a different quantile bin or drawing a different category) while keeping the rest of the row fixed,
\begin{equation}
\label{eq:typeII}
A'\sim P(A),\quad A'\neq A,\qquad R=(\bm{X}_{-A}, A', y),
\end{equation}
where $\bm{X}_{-A}$ denotes all features except $A$. We emphasize that Eq.~\ref{eq:typeII} is used as a \emph{regularizing penalty against dependency violations}, not as an estimator of the feature conditional $P(A\mid B)$: preferring $C$ over $R$ simply discourages unrealistic feature co-occurrences while leaving the learned joint $P(\bm{X})$ otherwise intact.

\paragraph{Sampling Strategy.}
To balance these objectives, we construct one rejected row for every training example. We select a Type I (Target) perturbation with probability $0.5$, and a Type II (Feature) perturbation otherwise. This ensures the model receives consistent feedback on the decision boundary $P(y\mid\bm{X})$ throughout training.

\subsection{Direct Preference Optimization for Tables}
\label{sec:dpo}

We fine-tune the generator using DPO~\cite{rafailov2023direct} to enforce the structural constraints defined above without an explicit reward model. 
In our tabular setting, the ``prompt'' is defined as the serialization of columns \emph{excluding} the perturbed variable, while the ``response'' is the perturbed variable itself. Equivalently, each row is decomposed into a fixed prompt $X$ and a completion $y$; $C$ and $R$ denote the chosen vs.\ rejected completions for the same prompt. To implement this with the random-order permutation of \textsc{GReaT}, we strictly enforce that the perturbed column appears \emph{after} its conditioning context (the other features) in the token sequence.

Let $\pi_{\mathrm{ref}}$ be the frozen base generator. We optimize the DPO-Positive objective~\cite{pal2024DPOP}:
\begin{align}
\mathcal L_{\mathrm{DPO}}(\pi_\theta;\pi_{\mathrm{ref}}) &=
-\mathbb E_{(X,C,R)\sim \mathcal{D}}
\!\left[
\log\sigma\!\bigl(
\beta\,\Delta(C,R\mid X)
\bigr)
\right] \nonumber\\
&\quad -\lambda\,
\mathbb E_{(X,C)\sim \mathcal{D}}
\left[
\max\!\Bigl(0,\,
\log\tfrac{\pi_{\mathrm{ref}}(C\mid X)}{\pi_\theta(C\mid X)}
\Bigr)
\right],
\end{align}
where $\Delta(C,R\mid X)=\log\frac{\pi_\theta(C\mid X)}{\pi_\theta(R\mid X)} - \log\frac{\pi_{\mathrm{ref}}(C\mid X)}{\pi_{\mathrm{ref}}(R\mid X)}$.
The first term increases the likelihood margin between the consistent row $C$ and the inconsistent row $R$, effectively pushing the synthetic distribution toward the true conditional dependencies. The second term is a regularization penalty that prevents the model from drifting too far from the valid data manifold learned during pre-training. We set $\beta{=}0.1$ and $\lambda{=}0.1$ in all experiments: $\beta$ is the standard DPO temperature~\cite{rafailov2023direct}, and $\lambda$ controls the strength of the DPO-Positive manifold-drift penalty~\cite{pal2024DPOP}. Both choices are robust---a sensitivity sweep (Appendix~\ref{sec:beta_lambda}) shows AUROC varies by $<0.004$ across $\beta\in\{0.05,0.1,0.2\}$, while removing the regularizer ($\lambda{=}0$) degrades AUROC by $0.069$, confirming it is necessary.

\subsection{Base Generator and Data Preparation}
\label{sec:preliminaries}

\paragraph{Data Augmentation.} 
Training deep generators on small tabular datasets ($N\lesssim10^3$) risks severe overfitting. To mitigate this, we first augment the training data via a SMOTE-like interpolation within categorical buckets. We synthesize additional rows by interpolating between nearest neighbors in the numeric subspace while preserving categorical signatures (see Appendix~\ref{sec:appendix_aug} for precise mixing ratios $M(N)$). This procedure enriches the manifold support without distorting conditional dependencies, providing a robust initialization for the generator. We emphasize that these augmented rows are used \emph{only} to pre-train the base generator $\pi_{\mathrm{ref}}$; they never appear in the final synthetic output or in any downstream evaluation, which always draws fresh rows from the trained generator and tests them on the held-out real data.

\paragraph{LM-based Architecture.}
We adopt the \textsc{GReaT} framework~\cite{borisov2022language} with a GPT-2 backbone. Data is serialized into natural language strings (e.g., \texttt{\small ``job is CEO, income is >=50k''}). To ensure the model learns the full joint distribution and remains permutation invariant, we randomly shuffle the feature order at every training step. Following \textsc{GReaT}, the joint likelihood then factorizes autoregressively over a random column permutation $\sigma$,
\begin{equation}
\label{eq:factorization}
P_\theta(x_1,\dots,x_d, y)=\textstyle\prod_{i} P_\theta\bigl(z_{\sigma(i)}\mid z_{\sigma(<i)}\bigr),
\end{equation}
where $z=(x_1,\dots,x_d,y)$. With limited data, optimizing all $d{+}1$ factors uniformly underweights the target factor $P_\theta(y\mid\bm{X})$; this is precisely the factor our DPO step (Section~\ref{sec:dpo}) selectively sharpens. The base generator $\pi_{\mathrm{ref}}$ is obtained by fine-tuning GPT-2 on this augmented textual data using standard autoregressive loss.

\section{Experiment}

In this section, we present the results of our experiments on real-world datasets. Our tests aim to address two key questions: (1) Does our reinforcement learning-based fine-tuning lead to superior statistical fidelity and downstream utility, under the realistic cases of small, imbalanced, and distribution-shifted data? (2) What are the effects of the main ReTabSyn components on data quality?

\begin{figure*}[t]
    \centering
    \begin{subfigure}{0.42\textwidth} 
        \centering
        \includegraphics[width=0.9\linewidth]{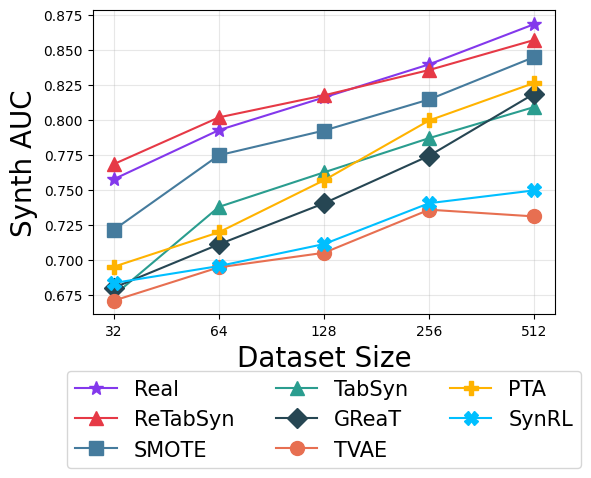}
        \caption{Synthetic data only}
        \label{fig:utility_panel1}
    \end{subfigure}
    \hspace{5mm} 
    \begin{subfigure}{0.42\textwidth} 
        \centering
        \includegraphics[width=0.9\linewidth]{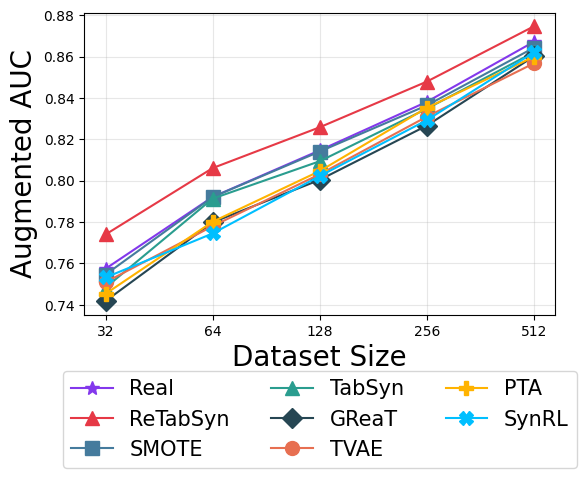}
        \caption{Augmentation on real data}
        \label{fig:utility_panel2}
    \end{subfigure}
    \caption{AUROC scores of synthetic data-trained models on in-distribution test sets, across varying training set sizes. The left panel shows performance on pure synthetic data, and the right panel on real data augmented with synthetic data.}
    \label{fig:utility_by_size}
\end{figure*}

\subsection{Test Setup}
\label{sec:preprocess}

\textbf{Datasets.} We evaluate on 10 benchmark tabular classification datasets~\cite{suh2023autodiff} with mixed numerical and categorical attributes (Table~\ref{tab:summary_statistics}; dataset details in Appendix~\ref{sec:dataset}). For categorical columns originally encoded as integer indices, we map indices to textual category names to better match our LM-based generator.

\begin{table}[]
    \centering
    \caption{Summary statistics of benchmark datasets}
    \label{tab:summary_statistics}
    \small
    \begin{tabular}{l r r r}
    \toprule
    \textbf{Dataset} & \textbf{Rows} & \textbf{Num Cols} & \textbf{Cate Cols} \\
    \midrule
    Adult                  & 48,812 & 6  & 8 \\
    Bean                   & 13,611 & 16 & 1 \\
    Churn-Modelling        & 10,000 & 6 & 4 \\
    HTRU2                  & 17,898 & 8  & 1 \\
    Indian Liver Patient   & 579    & 9  & 2 \\
    Obesity                & 2,111  & 8  & 9 \\
    Shoppers               & 12,330 & 16 & 2 \\
    Magic                  & 19,000 & 10 & 1 \\
    Titanic                & 714    & 6  & 2 \\
    Wilt                   & 4,839  & 5  & 1 \\
    \bottomrule
    \end{tabular}
\end{table}

\textbf{Scenarios and protocol.} Each dataset uses a fixed 80/20 train/test split. We evaluate three realistic regimes: (i)~\emph{Small Data}: train on random subsets of $N \in \{32, 64, 128, 256, 512\}$ rows; (ii)~\emph{Imbalanced Data}: for binary targets, down-sample the minority class to 1\% prevalence in both train and test; (iii)~\emph{Distribution Shift}: train/test are formed by demographic splits and the split column is removed (Appendix~\ref{sec:shift}). We repeat all settings over 10 random seeds and sample 1024 synthetic rows per generator.

\textbf{Baselines and metrics.} Baselines include SMOTE~\cite{chawla2002smote}, TVAE~\cite{xu2019modeling}, TabSyn~\cite{zhang2023tabsyn}, GReaT~\cite{borisov2022language}, and the RL-guided methods PTA and SynRL~\cite{yang2024pta, das2024synrl}; training details follow official implementations (Appendix~\ref{sec:baseline}). Utility is measured by training logistic regression, Naive Bayes, decision tree, random forest, XGBoost~\cite{chen2016xgboost}, and CatBoost~\cite{prokhorenkova2018catboost} on synthetic data and evaluating on the fixed real test set; we report mean AUROC (or PR-AUC for the 1\% prevalence setting). We also report fidelity ($\alpha$-Precision/$\beta$-Recall, marginal similarity, correlation similarity) and privacy (membership-inference Leakage and Authenticity); definitions are in Appendix~\ref{sec:privacy_impl}.

\subsection{Small Data Test}\label{sec:smalldata}



Figure~\ref{fig:utility_by_size} shows the AUROC scores of classifiers trained on synthetic data generated by various models, as a function of the number of training rows. In the low-data regime (32--128 rows), our proposed method, ReTabSyn, consistently outperforms all deep generator baselines, including GReaT, TVAE, TabSyn and RL-based methods PTA and SynRL. Remarkably, when training data is extremely limited, ReTabSyn even surpasses the performance achieved with real data, likely due to the advantage conferred by the large volume of synthetic samples.

As the size of the training set increases, the overall utility of synthetic data improves for all generators; however, ReTabSyn maintains its lead, consistently delivering the highest AUROC scores. The advantage is most pronounced in the smallest regimes and is partly obscured by averaging across sizes: in the synthetic-only setting, ReTabSyn exceeds the strong SMOTE baseline by $+0.056$, $+0.018$, and $+0.021$ AUROC at $N=32$, $64$, and $128$, respectively, with the gap narrowing as $N$ grows (full per-size mean$\pm$SE in Appendix~\ref{sec:smalldata_se}). When synthetic data is used to augment the real training set, the performance gap between ReTabSyn and the baselines becomes even more pronounced in the low-data regime. Only at larger training sizes does the interpolation-based SMOTE method close the gap to our approach, which we attribute to a shifted trade-off between utility and data privacy in interpolation-based technique (more discussion in \ref{sec:fidelity}).


\subsection{Imbalanced Target}\label{sec:imbalanced}



\begin{table*}[t]
\centering
\caption{PR-AUC at 1\% prevalence (mean$_{\pm\mathrm{SE}}$ over 10 seeds). Best synthetic generator per dataset in \textbf{bold} (Real excluded).}
\small
\begin{tabular}{lccccc}
\toprule
Model       & Adult & Churn & HTRU2 & Magic & Shoppers \\
\midrule
Real        & \se{0.924}{.009} & \se{0.883}{.014} & \se{0.897}{.008} & \se{0.916}{.012} & \se{0.701}{.013} \\ \hline
TVAE        & \se{0.861}{.015} & \se{0.821}{.013} & \se{0.851}{.012} & \se{0.861}{.011} & \se{0.590}{.012} \\
GReaT       & \se{0.873}{.015} & \se{0.834}{.010} & \se{0.857}{.014} & \se{0.873}{.009} & \se{0.604}{.011} \\
SMOTE       & \se{0.891}{.010} & \se{0.852}{.016} & \se{0.869}{.013} & \se{0.889}{.015} & \se{0.649}{.009} \\
TabSyn      & \se{0.882}{.017} & \se{0.843}{.011} & \se{0.862}{.010} & \se{0.882}{.012} & \se{0.632}{.015} \\
SynRL       & \se{0.862}{.014} & \se{0.825}{.012} & \se{0.852}{.011} & \se{0.863}{.011} & \se{0.635}{.011} \\
PTA         & \se{0.871}{.013} & \se{0.832}{.011} & \se{0.856}{.011} & \se{0.872}{.010} & \se{0.623}{.010} \\
\textbf{ReTabSyn} & \seb{0.906}{.013} & \seb{0.866}{.014} & \seb{0.881}{.012} & \seb{0.901}{.010} & \seb{0.670}{.012} \\
\bottomrule
\end{tabular}
\label{tab:prauc_imbalanced}
\end{table*}

\textbf{Dataset eligibility.} This setting requires a \emph{binary} target with enough effective support to construct a 1\% minority-prevalence split while retaining a stable train/validation/test evaluation (at least ${\sim}20$ minority examples per evaluation split). Datasets failing this criterion (e.g., the multi-class Bean/Obesity, or datasets with insufficient positives at 1\%) are not applicable rather than selectively omitted; a full per-dataset eligibility audit is given in Appendix~\ref{sec:eligibility}. Under this rule we additionally include \textbf{Shoppers}, which was inadvertently omitted in an earlier draft.

Table~\ref{tab:prauc_imbalanced} reports PR-AUC at 1\% prevalence. Across all datasets, ReTabSyn is closest to the real-data benchmark and outperforms every generator baseline, indicating better preservation of minority-class signal under extreme imbalance. For example, on Adult ReTabSyn reaches 0.906 PR-AUC versus 0.891 for the next-best baseline (SMOTE), with non-overlapping standard-error intervals.


\subsection{Shifted Distribution Robustness}\label{sec:shifted}

\begin{table*}[t]
\centering
\caption{AUC under distribution shift (mean$_{\pm\mathrm{SE}}$ over 10 seeds). Best synthetic generator per dataset in \textbf{bold} (Real excluded). }
\small
\begin{tabular}{lcccc}
\toprule
Model       & Adult & Churn & Liver & Titanic \\
\midrule
Real        & \se{0.868}{.011} & \se{0.720}{.012} & \se{0.603}{.013} & \se{0.862}{.012} \\ \hline
TVAE        & \se{0.806}{.014} & \se{0.686}{.014} & \se{0.565}{.015} & \se{0.823}{.013} \\
GReaT       & \se{0.819}{.013} & \se{0.701}{.014} & \se{0.576}{.015} & \se{0.836}{.012} \\
SMOTE       & \se{0.817}{.012} & \se{0.704}{.013} & \se{0.596}{.013} & \se{0.851}{.011} \\
TabSyn      & \se{0.808}{.015} & \se{0.704}{.016} & \se{0.594}{.016} & \se{0.849}{.014} \\
SynRL       & \se{0.809}{.013} & \se{0.688}{.013} & \se{0.565}{.014} & \se{0.825}{.012} \\
PTA         & \se{0.821}{.012} & \se{0.702}{.012} & \se{0.577}{.013} & \se{0.837}{.011} \\
\textbf{ReTabSyn} & \seb{0.863}{.011} & \seb{0.721}{.012} & \seb{0.618}{.013} & \seb{0.860}{.011} \\
\bottomrule
\end{tabular}
\label{tab:auc_shift}
\end{table*}

\textbf{Dataset eligibility.} This setting requires a semantically meaningful, reproducible split column (a categorical demographic/geographic attribute with a majority:minority ratio $\geq 1.5{:}1$) that induces a genuine covariate shift; datasets lacking such a column (e.g., HTRU2, Magic) cannot form a meaningful shift benchmark and are excluded (full audit in Appendix~\ref{sec:eligibility}).

Table~\ref{tab:auc_shift} reports AUC on distribution-shifted splits (Adult, Churn, Liver, Titanic), where models are trained on a majority subgroup and evaluated on a minority subgroup (§\ref{sec:preprocess}).
Across all datasets, \textbf{ReTabSyn} is consistently closest to the real-data upper bound and outperforms every synthetic baseline, including the RL-guided methods that rely on external classifiers (PTA, SynRL).  
SMOTE, TabSyn, and GReaT trail ReTabSyn by 1–4 AUC points, highlighting that simple interpolation or unconditional generation offers limited protection against covariate shift.  
These results indicate that our oracle-free alignment better preserves feature–target dependencies that transfer across subpopulations.

\begin{figure*}[!ht]
    \centering
    \includegraphics[width=0.95\textwidth]{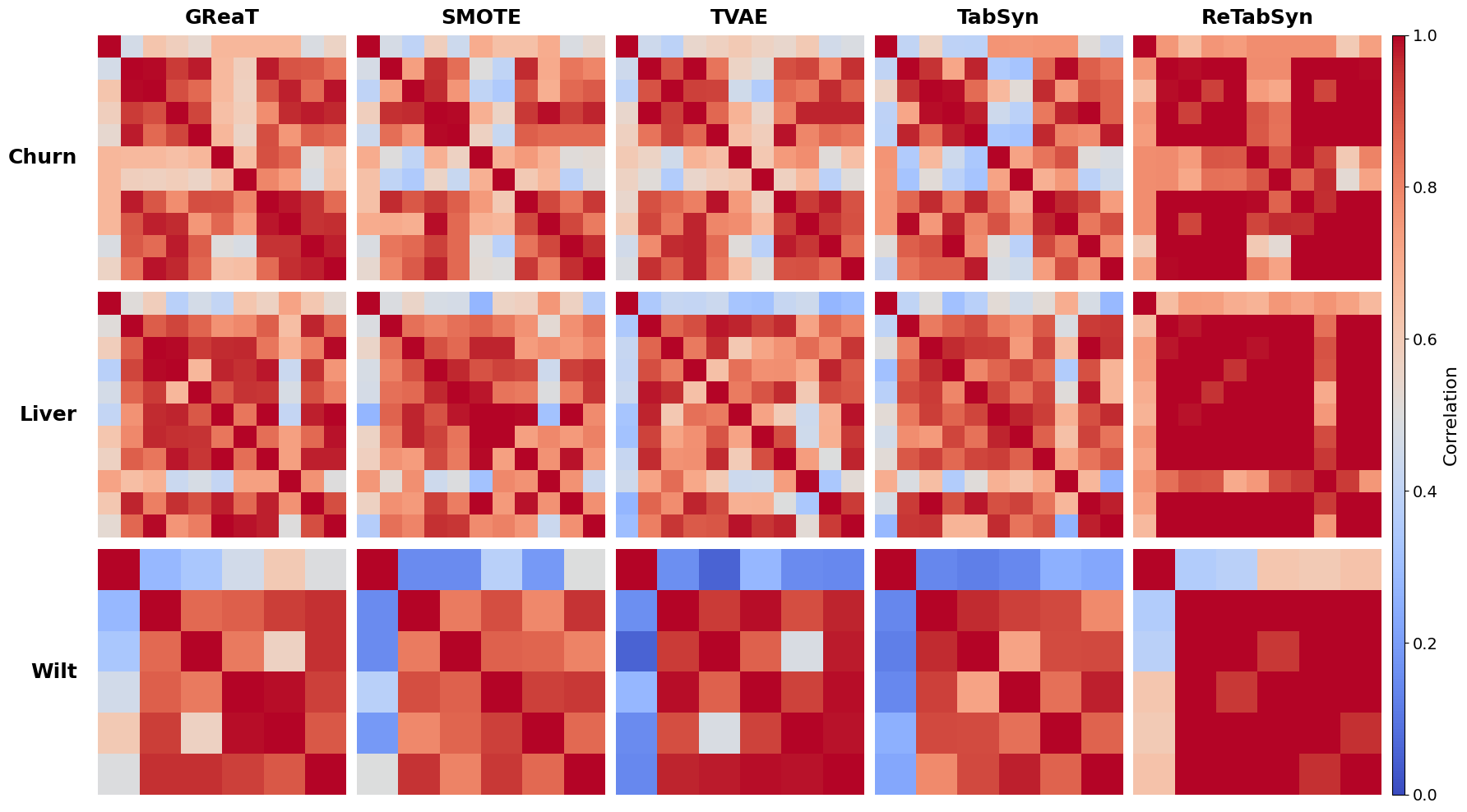}
    \caption{Correlation similarity matrix visualized using a color scale. Shades closer to red indicate high similarity, meaning the correlation between real and synthetic data is well preserved. In contrast, shades of blue signify low similarity, suggesting a weaker alignment between real and synthetic feature correlation.}
    \label{fig:correlation_similarity}
\end{figure*}

\subsection{Statistical Fidelity}
\label{sec:fidelity}

\begin{table}
\centering
\caption{Fidelity metrics across models (mean$_{\pm\mathrm{SE}}$ over 10 seeds; averaged over all small-data splits). Best per column in \textbf{bold}. Shape is the average marginal-distribution similarity (1–KS/TVD) and Corr.\ is the mean pairwise correlation similarity.}
\small
\setlength{\tabcolsep}{4pt}
\begin{tabular}{lcccc}
\toprule
Model & Precision & Recall & Shape & Corr. \\
\midrule
TVAE & \se{0.280}{.016} & \se{0.180}{.010} & \se{0.760}{.006} & \se{0.760}{.005} \\
SMOTE & \seb{0.530}{.011} & \se{0.320}{.008} & \seb{0.830}{.005} & \se{0.810}{.003} \\
TabSyn & \se{0.410}{.019} & \se{0.290}{.010} & \se{0.810}{.005} & \se{0.800}{.004} \\
GReaT & \se{0.440}{.017} & \se{0.210}{.010} & \se{0.800}{.005} & \se{0.740}{.003} \\
SynRL & \se{0.300}{.015} & \se{0.200}{.009} & \se{0.780}{.005} & \se{0.780}{.005} \\
PTA & \se{0.430}{.017} & \se{0.220}{.010} & \se{0.790}{.005} & \se{0.750}{.006} \\
ReTabSyn & \se{0.510}{.017} & \seb{0.330}{.010} & \seb{0.830}{.005} & \seb{0.830}{.004} \\
\bottomrule
\end{tabular}
\label{tab:Diversity Metrics}
\end{table}
We next examine \emph{statistical fidelity}: the extent to which synthetic samples reproduce the density and dependence structure of the real data, because these properties underpin downstream utility.  
Table~\ref{tab:Diversity Metrics} summarizes four fidelity metrics averaged over all small data splits in Section~\ref{sec:preprocess}.  

ReTabSyn attains the highest (or tied-highest) scores on \textit{$\beta$-Recall}, \textit{Shape}, and \textit{Corr.}, indicating that its samples cover more of the real data’s high-density regions while preserving inter-feature structure.  
SMOTE achieves the best \textit{$\alpha$-Precision}, reflecting its interpolation of real points, yet it incurs markedly higher membership-inference leakage (AUC 0.83 versus 0.62 for ReTabSyn; see Table~\ref{tab:privacy_attack}), highlighting the privacy risk of copy-based generators.
\textit{PTA} and \textit{SynRL}—RL-guided methods that rely on external classifiers—do not improve fidelity: both trail ReTabSyn on \textit{Recall}, \textit{Shape}, and \textit{Corr.} (with SynRL also lowest on \textit{Precision}), suggesting that classifier-driven signals alone are insufficient to model the joint structure needed for high-fidelity synthesis.
ReTabSyn’s authenticity is slightly below that of its base model (GReaT), which we attribute to the extra fine-tuning steps that increase memorization, but it still offers a favorable trade-off between fidelity and privacy.  
Overall, the balanced performance across metrics confirms ReTabSyn’s ability to generate synthetic data that is both faithful and diverse.


Figure~\ref{fig:correlation_similarity} visualizes pairwise-correlation similarity for three representative datasets (darker red = higher).  
ReTabSyn retains the strongest block-diagonal patterns, whereas GReaT, TabSyn, and TVAE show progressively lighter cells, signaling weakened or inverted dependencies.  
By better matching these correlations within \(X\) and between \(X\) and \(y\), ReTabSyn produces a conditional distribution \(\tilde P(y\mid X)\) that is closer to the real one, which in turn translates into superior downstream performance.  
These findings reinforce the value of our reinforcement-learning alignment strategy for enforcing critical statistical relationships and delivering reliable, high-utility synthetic data.

\subsection{Privacy Risk}

\begin{table}
\caption{Privacy attack metrics over 10 seeds. Leakage AUC and TPR@FPR=0.01 are the empirical worst-case (maximum over 13 MIAs), lower-is-better; Authenticity is higher-is-better. Values reported as mean$_{\pm\mathrm{SE}}$ where available.}
\centering
\small
\begin{tabular}{lccc}
\toprule
Model       & AUC    & TPR@FPR=0.01   & Authenticity \\
\midrule
TVAE        & \se{0.627}{.011} & \se{0.061}{.006} & \se{0.743}{.008} \\
SMOTE       & \se{0.831}{.009} & \se{0.438}{.004} & \se{0.635}{.006} \\
TabSyn      & \se{0.600}{.011} & \se{0.052}{.005} & \se{0.671}{.011} \\
GReaT       & \se{0.587}{.010} & \se{0.054}{.005} & \se{0.650}{.009} \\
SynRL       & \se{0.625}{.008} & \se{0.060}{.004} & \se{0.740}{.008} \\
PTA         & \se{0.589}{.012} & \se{0.055}{.005} & \se{0.652}{.009} \\
ReTabSyn    & \se{0.620}{.011} & \se{0.060}{.003} & \se{0.640}{.009} \\
\bottomrule
\end{tabular}
\label{tab:privacy_attack}
\end{table}

Table~\ref{tab:privacy_attack} reports worst-case membership-inference results (AUC, TPR at 1\% FPR) and \emph{Authenticity} (higher is better).  
Relative to the interpolation baseline \textbf{SMOTE}, \textbf{ReTabSyn} substantially reduces leakage (AUC 0.62 vs.\ 0.83; TPR 0.06 vs.\ 0.44) while achieving comparable Authenticity.  
Against other deep generators, including the RL-guided, classifier-driven \textit{PTA} and \textit{SynRL}—ReTabSyn’s leakage is in the same range (AUC $\approx$ 0.59–0.63; TPR $\approx$ 0.05–0.06).  
Its Authenticity (0.64) is close to GReaT/PTA and slightly below SynRL, reflecting our additional fine-tuning, yet ReTabSyn delivers stronger \emph{utility} (Secs.~\ref{sec:smalldata}–\ref{sec:shifted}).  
Overall, ReTabSyn occupies a favorable privacy–utility trade-off: it maintains table realism on par with strong baselines while exposing markedly less information than interpolation and no more than competing deep generators.

\subsection{Ablation Study of Design Choices}
\label{sec:ablation}

\begin{table}[t]
    \centering
    \caption{Ablation on the Wilt dataset. In each block, a single parameter or settings is varied while others remain at their default values. Metrics include AUROC on synthetic data trained classifiers, correlation similarity, precision, and recall.}
    \small
    \renewcommand{\arraystretch}{0.95}
    \begin{tabular}{p{2.8cm}cccc}
        \toprule
         & AUROC & Corr. & Prec. & Recall \\
        \midrule \midrule
        \textbf{Default ReTabSyn}  & 0.841 & 0.880 & 0.763 & 0.471 \\
        \midrule
        \multicolumn{5}{l}{\textbf{Training stages}} \\
        Only DPO (no aug.)   & 0.800 & 0.810 & 0.713 & 0.429 \\
        Only Aug. pre-train   & 0.815 & 0.851 & 0.749 & 0.420 \\
        \midrule
        \multicolumn{5}{l}{\textbf{Reward supervision (§\ref{sec:preference})}} \\
        Oracle editing   & 0.809 & 0.867 & 0.731 & 0.427 \\
        Oracle screening & 0.817 & 0.840 & 0.726 & 0.385 \\
        \midrule
        \multicolumn{5}{l}{\textbf{Preference pairs with target (§\ref{sec:dpo})}} \\
        Proportion = 0   & 0.821 & 0.888 & 0.767 & 0.493 \\
        Proportion = 1   & 0.843 & 0.862 & 0.755 & 0.448 \\
        \midrule
        \multicolumn{5}{l}{\textbf{Augmentation size (§\ref{sec:preliminaries})}} \\
        Nx1   & 0.827 & 0.836 & 0.748 & 0.426 \\
        Nx20  & 0.837 & 0.878 & 0.759 & 0.460 \\
        1000 (fixed)  & 0.835 & 0.839 & 0.750 & 0.435 \\
        \bottomrule
    \end{tabular}
    \label{tab:factor_analysis}
\end{table}

Table~\ref{tab:factor_analysis} ablates ReTabSyn on Wilt, linking each setting to its corresponding methodology component. For the within-bucket augmentation (§\ref{sec:preliminaries}), fixing augmentation size underperforms the dynamic schedule, while very large augmentation (Nx20) yields diminishing returns; the dynamic strategy adapts to dataset size, avoiding both data starvation and overfitting. Removing either training stage, skipping augmented pre-training or skipping DPO fine-tuning, reduces both AUROC and fidelity, confirming that the two-stage pipeline is necessary: augmentation provides diverse feature relationships, while DPO sharpens the conditional distribution. For the preference labeling procedure (§\ref{sec:preference}), we compare our oracle-free perturbation approach against two oracle-based variants: \emph{editing} (using a trained classifier to correct target values) and \emph{screening} (rejecting rows whose predictions disagree); both underperform the oracle-free default, suggesting that oracle noise degrades preference quality. Finally, varying the fraction of preference pairs that include the target column $y$ (§\ref{sec:dpo}) reveals a trade-off: all feature--feature pairs improve correlation/recall but lower AUROC, while all target pairs boost AUROC but reduce correlation; the default 50/50 mix provides the best overall balance between utility and fidelity.

\begin{figure}[t]
    \centering
    \includegraphics[width=0.9\columnwidth]{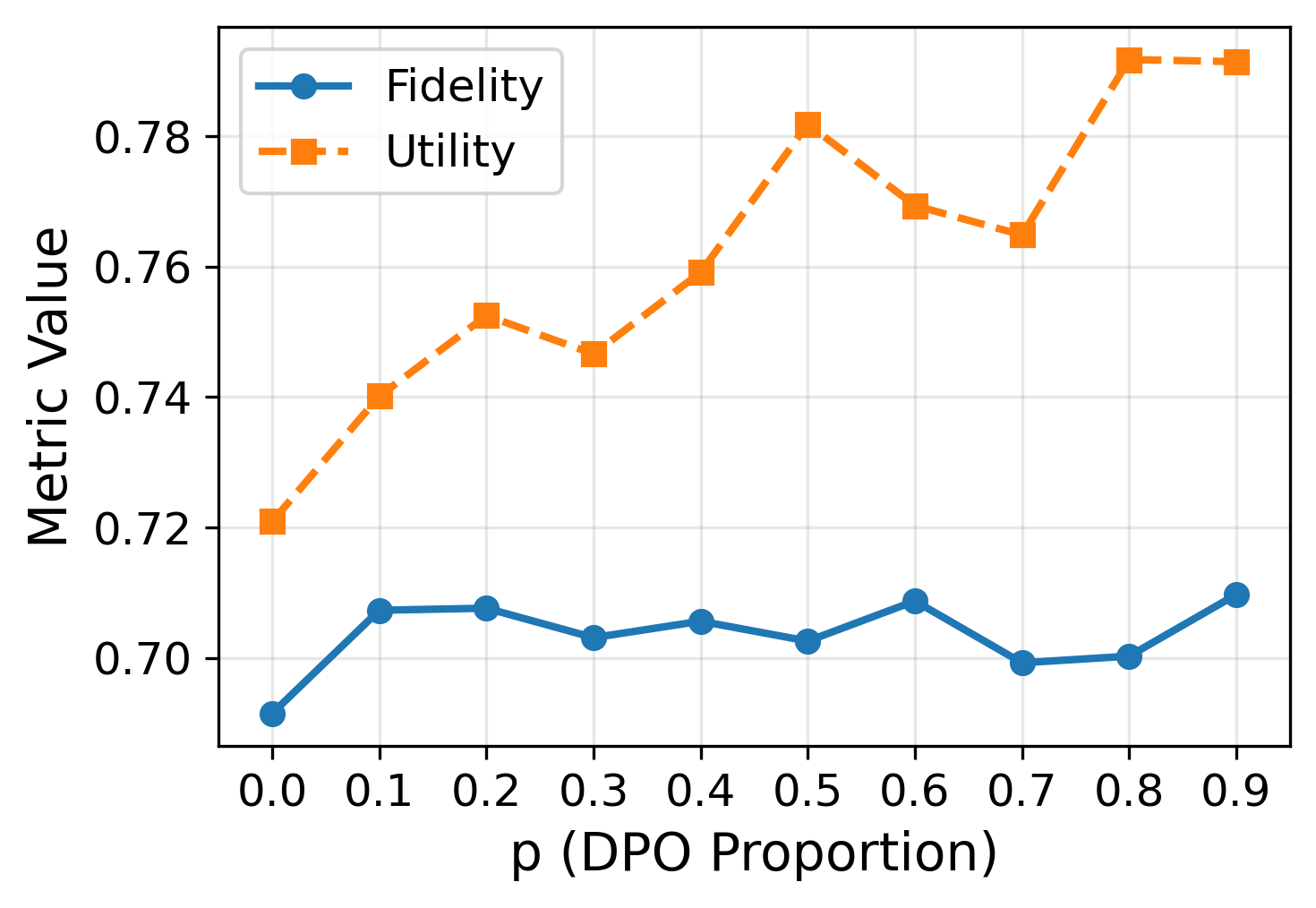}
    \caption{Effect of varying the proportion $\rho$ of DPO steps relative to total fine-tuning steps. As $\rho$ increases, fidelity remains stable while utility (ROC AUC) consistently improves.}
    \label{fig:cond_priority}
\end{figure}

We also vary the fraction $\rho$ of the fixed fine-tuning budget allocated to DPO, performing SFT first to learn the joint $P(\bm{X}, y)$ and then DPO to sharpen the conditional $P(y\mid\mathbf{X})$ (Figure~\ref{fig:cond_priority}). As $\rho$ increases, fidelity stays stable while utility (ROC AUC) consistently improves, confirming that prioritizing the conditional distribution helps most in low-data regimes, consistent with our motivation in §\ref{sec:theory}.




We also evaluate reality constraints on Adult, testing three categorical rules that encode domain knowledge (e.g., ``marital-status = Widow $\Rightarrow$ sex = Female''). ReTabSyn achieves 0.00\% violations across all rules, compared to 1--3\% for baselines (see Appendix~\ref{sec:constraint_appendix}, Table~\ref{tab:constraints_appendix}). This stems from our preference-pair construction, which perturbs constraint-related variables so DPO penalizes dependency-breaking rows.

Finally, ReTabSyn's gains generalize beyond its GPT-2 backbone and its trained target. On a stronger backbone (Qwen3.5-2B) it lifts synthetic-only AUROC to 0.900/0.858 on Wilt/Churn, above SMOTE and a prompted frontier model (GPT-5.4; Appendix~\ref{sec:backbone}); across 12 alternative-target evaluations it attains the highest mean AUROC, winning 8/12 (Appendix~\ref{sec:multitarget}); and on full balanced data (Adult, Magic) it stays competitive with SMOTE while consistently beating its GReaT base (Appendix~\ref{sec:fulldata}).

\section{Conclusion}

We introduced \textbf{ReTabSyn}, a reinforcement learning–aligned tabular synthesizer that uses DPO on oracle-free target and feature perturbations to directly preserve feature–target dependencies. By prioritizing synthetic examples that better match \(P(y\mid \mathbf X)\), ReTabSyn consistently improves downstream ML utility. Across low-data, class-imbalance, and distribution-shift settings, it outperforms strong generators—including GReaT, TVAE, TabSyn, and the classifier-guided RL baselines PTA and SynRL—while also reducing structural-rule violations.  Beyond accuracy, fidelity-aware alignment provides a practical path to \emph{controllable} synthesis: enforcing domain constraints and steering distributions in ways that can support fairness and privacy goals. We see this as a step toward robust, utility-preserving synthetic data for real-world use.

\paragraph{Limitations.}
ReTabSyn has several limitations. (i)~Its gains are largest in low-data, imbalanced, and shifted regimes; in full, balanced data the margin over strong interpolation baselines such as SMOTE narrows (Appendix~\ref{sec:fulldata}). (ii)~It is stronger on classification than on \emph{regression}, where it trails SMOTE, since our perturbation design is more naturally matched to discrete targets (Appendix~\ref{sec:regression}). (iii)~The DPO fine-tuning slightly lowers authenticity relative to the base generator (Section~\ref{sec:fidelity}), though leakage remains far below interpolation methods. (iv)~ReTabSyn is target-aware, using one chosen alignment attribute during preference construction; while its benefits transfer to other targets (Appendix~\ref{sec:multitarget}), it is not fully task-agnostic. (v)~The two-stage pipeline (augmented pre-training plus DPO) adds modest compute over single-stage generators, although the DPO step itself is lightweight (Appendix~\ref{sec:runtime-decomposition}).

For future work, we plan to (i) test ReTabSyn with alternative backbones (e.g., diffusion) to assess generality and scaling; (ii) design regression-aware preference signals for continuous targets; (iii) incorporate explicit constraints for fairness, privacy, and adversarial robustness; and (iv) study principled controls to mitigate bias and support ethical data practices.

\section*{Acknowledgements}
This work was supported in part by a gift from Cisco, the National Science Foundation (NSF CNS-2247795), and the Office of Naval Research (ONR N00014-22-1-2680).

\section*{Impact Statement}
This paper proposes a novel method for generating synthetic tabular data with improved downstream utility under realistic low-data, class-imbalanced, and distribution-shift settings. This method can enable sharing and using sensitive structured datasets by enabling privacy-preserving data release and safer model development workflows, which may benefit domains such as healthcare, finance, and the social sciences.

At the same time, synthetic data generation poses risks, such as leaking information about individuals in the training data if the generator memorizes rare or unique records, amplifying existing societal biases present in the source data, and creating synthetic datasets that appear credible but support deceptive analyses.

We explicitly evaluate privacy-related risks using membership-inference-style auditing and report utility--privacy trade-offs. Nevertheless, privacy and fairness guarantees are out of the scope of this paper. We recommend performing privacy audits and distributional diagnostics prior to release, and avoiding deployment in high-stakes decision settings without domain-specific validation and fairness assessment.

\bibliography{kdd}
\bibliographystyle{icml2026}

\newpage
\appendix
\onecolumn

\Huge  \textbf{Appendix}
\normalsize 

\section{Dataset}
\label{sec:dataset}
We provide the URL for the sources of each downstream benchmark set considered in the paper. 

\begin{enumerate}
    \item \textbf{Adult} (Kohavi, R): ~\cite{kohavi1996scaling}. (Binary class)
    \item \textbf{Bean} (UCI): \href{https://archive.ics.uci.edu/dataset/602/dry+bean+dataset}{Link} (Multi class)
    \item \textbf{Churn} (UCI): \href{https://www.kaggle.com/datasets/shrutimechlearn/churn-modelling}{Link} (Binary class)
    \item \textbf{HTRU2} (UCI): \href{https://archive.ics.uci.edu/dataset/372/htru2}{Link} (Binary class)
    \item \textbf{Indian Liver Patient} (Kaggle): \href{https://www.kaggle.com/datasets/uciml/indian-liver-patient-records?resource=download}{Link} (Binary class)
    \item \textbf{Obesity} (Kaggle): \href{https://www.kaggle.com/datasets/tathagatbanerjee/obesity-dataset-uci-ml}{Link} (Multi class)
    \item \textbf{Shoppers} (Kaggle): \href{https://www.kaggle.com/datasets/henrysue/online-shoppers-intention}{Link} (Binary class)
    \item \textbf{Magic} (Kaggle): \href{https://www.kaggle.com/datasets/abhinand05/magic-gamma-telescope-dataset?resource=download}{Link} (Binary class)
    \item \textbf{Titanic} (Kaggle): \href{https://www.kaggle.com/c/titanic/data}{Link} (Binary class)
    \item \textbf{Wilt} (OpenML): \href{https://www.openml.org/search?type=data\&sort=runs\&id=40983\&status=active}{Link} (Binary class)
\end{enumerate}

\section{Training Data Augmentation}
\label{sec:appendix_aug}

A key challenge when fine-tuning a table generator on small datasets ($N\lesssim10^3$) is that many category–numeric combinations occur only once, leading the model to memorize individual rows. To mitigate this, we synthesize $M(N) * N$ additional rows via a SMOTE-like interpolation within each categorical bucket, where
\[
M(N)=
\begin{cases}
30,&N\le128,\\
10,&128<N\le256,\\
5,&256<N\le1000,\\
1,&N>1000,
\end{cases}
\]
so that augmentation is aggressive for very small tables and diminishes as $N$ grows. Concretely, we first drop any rows with missing values, then group the real data by the full signature of its categorical columns. For each synthetic row, we sample a bucket with probability proportional to its size, fit a $k$-NN index ($k=5$) on the bucket’s numeric submatrix, pick a random seed example $x_s$ and one of its neighbors $x_n$, draw $\lambda\sim\mathrm{Uniform}(0,1)$, and set $\tilde x = x_s + \lambda(x_n - x_s)$. We round any integer-typed columns to the nearest integer and retain the original category values. This procedure preserves realistic feature–category dependencies, enriches sparse regions of the numeric manifold, and keeps the total pre-training size bounded by $(1+M(N))N\le31N$, ensuring both improved diversity and efficient training. The resulting smoother local neighborhood structure is later exploited when constructing minimally invasive perturbations for preference labeling.

\section{Distribution Shifted Dataset}
\label{sec:shift}

We detail the splitting scheme for creating distribution shifted example in table~\ref{tab:demographic_shift}. Once the train and test split is created, the split column is dropped from both train and test to avoid mismatch of feature category sets.

\begin{table*}[!ht]
    \centering
    \begin{tabular}{l|l|l|r|r}
        \hline
        \textbf{Dataset Name} & \textbf{Split Column} & \textbf{Train / Test Composition} & \textbf{N Train} & \textbf{N Test} \\
        \hline
        \textbf{Adult} & native-country & United States / non-US & 43,152 & 5,128 \\
        \textbf{Indian liver patient} & Gender & Male / Female & 441 & 142 \\
        \textbf{Obesity} & Gender & Male / Female & 1068 & 1043 \\
        \textbf{Churn Modeling} & Country & France \& Germany / Spain & 7,519 & 2,477 \\
        \textbf{Titanic} & Sex & Male / Female & 577 & 314 \\
        \hline
    \end{tabular}
    \caption{Dataset splits based on demographic variables to induce distributional drift.}
    \label{tab:demographic_shift}
\end{table*}

\section{Baseline Implementation}
\label{sec:baseline}
\textbf{GReaT~\cite{borisov2022language}:} We used the official GitHub implementation. We used a batch size of 32. During pre-training, we began with a pre-trained distilgpt2 model and trained for 2 million steps on the combination of pre-training data. We train 200 epochs for each dataset during fine-tuning.

\textbf{TabSyn~\cite{zhang2023tabsyn}:} We used the official GitHub implementation with default parameters. For pre-training with heterogeneous VAE embeddings, we train its VAE model for each pre-training dataset, zero-pad all embeddings to the same dimension, and then pre-train a diffusion model on such padded embeddings. During downstream training, the VAE embedding of the downstream datasets are padded to the same dimension as in the pre-training. The pre-trained TabSyn is loaded and diffusion training proceed with it as initialization.

\textbf{SMOTE~\cite{chawla2002smote}:} The original SMOTE algorithm is designed to upsample minority classes. We extend it to perform interpolation for all classes following the implementation from~\cite{kotelnikov2023tabddpm}. For each generation, we first randomly select one target class using empirical class frequency as probability. Then we randomly sample one example from the selected class, and generated interpolated examples using number of nearest neighbour $k=5$. The interpolation weight $\alpha = 0.5$.

\textbf{TVAE~\cite{xu2019modeling}}: We used the official SDV library implementation. We used default parameters: class dimensions =(256, 256, 256, 256), random dimensions=100, 64 channels, l2scale=1e-5, batch size=500, training epoch = 300.

\textbf{PTA~\cite{yang2024pta}}: We used the official GitHub implementation. We use the default GPT-2 as the base language model with total training epochs of 3. The classifier training employs batch size of 2, learning rate of $2 \times 10^{-5}$, and N=2 samples per class. The GAN component uses hidden dimensions of (256, 256), batch size of 32, and trains for 100 epochs with dropout rate of 0.2. The method processes data in chunks of 1000 samples and applies gradient clipping with maximum norm of 1.0.

\textbf{SynRL~\cite{das2024synrl}}: We used the official implementation. We followed the default setting with embedding dimension of 128, hidden layer dimensions of (128, 128), and batch size of 500. The model uses Adam optimizer with weight decay of $1 \times 10^{-5}$ and trains for 300 epochs. The Proximal Policy Optimization (PPO) component operates with learning rate of $1 \times 10^{-4}$, clipping range of 0.01, and batch size of 50. Discrete column detection uses maximum unique ratio threshold of 0.05 and maximum unique count of 20 for automatic categorical variable identification.






\section{Privacy Implementation}
\label{sec:privacy_impl}

\subsubsection{Privacy Leakage}

\paragraph{Threat model.}
We consider black-box and model-unknown shadow-box
adversaries who see the released synthetic set \(S\)—
optionally supplemented by a reference set \(R\)—
but never the generator code or parameters.
This mirrors realistic data-sharing scenarios and enables model-agnostic,
computationally feasible audits~\cite{vanbreugel2023membership,golob2024privacyvulnerabilitiesmarginalsbasedsynthetic}.

\paragraph{Auditing procedure.}
Following the empirical-worst-case principle of
Empirical Differential Privacy (EDP)~\cite{Jagielski2020},
we measure leakage as the \emph{maximum} attack AUC across
\(\mathcal{A}=13\) state-of-the-art MIAs that span distance, density,
and classifier signals:
\[
\text{Leakage} \;=\;
\max_{A\in\mathcal{A}} \text{AUC}(A).
\]
The Area Under Receiver Operating Characteristic curve (ROC AUC) and True Positive Rate (TPR) at False Positive Rate (FPR) are used as performance metrics of attacks, where a higher performance indicates more success in identifying membership of training data points, and thus greater privacy leakage. All attacks are re-implemented in a common Python framework; no model
re-training is required.  Table~\ref{tab:mia_list} lists the methods. We used the Synth-MIA~\cite{ward2025synthmia} library as our attack framework implementation. 

\begin{table}[ht]
\centering
\caption{Membership-inference attacks used in this study.}
\label{tab:mia_list}
\begin{tabular}{lll}
\toprule
\textbf{Attack} & \textbf{Threat model} & \textbf{Signal type} \\
\midrule
DOMIAS \cite{vanbreugel2023membership}           & Shadow-box & Density ratio \\
DPI \cite{ward2024dataplagiarismindexcharacterizing}              & Shadow-box & Local density \\
Classifier \cite{houssiau2022tapas}      & Shadow-box & Density ratio \\
Density Estimator \cite{houssiau2022tapas} & Black-box & Density estimation \\
DCR \cite{ganleaks}                     & Black-box & Distance-based \\
DCR-Diff \cite{ganleaks}                & Shadow-box & Distance difference \\
Logan \cite{Hayes2017LOGANMI}            & Shadow-box & Density ratio \\
MC Estimation \cite{Hilprecht2019MonteCA} & Black-box & Density estimation \\
\bottomrule
\end{tabular}

\end{table}

\section{Computation and Run‑time Decomposition}
\label{sec:runtime-decomposition}

Our RL fine-tuning experiments are conducted using the Hugging Face \texttt{trl} library with a pre-trained GPT-2 causal language model as the backbone. For DPO training, we set \(\beta = 0.1\) and use an initial learning rate of \(1 \times 10^{-4}\) over 3 training epochs. Both ReTabSyn and baseline experiments are run on a server equipped with a single NVIDIA H100 GPU (96 GB GPU memory) and 128 GB of RAM, using a batch size of 64 for both SFT and DPO training.

\begin{table}[ht]
  \centering
  \caption{Run‑time decomposition of generator operations (Train, Fine‑tune, Sampling) on full Wilt dataset}
  \label{tab:runtime-decomposition}
  \begin{tabular}{lccc}
    \toprule
    Model & Train & Fine‑tune & Sampling \\
    \midrule
    ReTabSyn & 1hrs & 3mins & 7mins \\
    GReaT & 1hrs & - & 7mins \\
    TabSyn & 55mins & - & 2mins \\
    SMOTE & - & - & 3mins \\
    \bottomrule
  \end{tabular}
\end{table}

Table~\ref{tab:runtime-decomposition} shows that the majority of elapsed time is spent on training the base generator—especially for large, LM‑based models—while the RL‑based fine‑tuning step in ReTabSyn takes only a few minutes. In practice, this means that although the overall workflow may appear time‑intensive due to the initial model training, the preference‑optimization phase is lightweight and can be applied to any pre‑trained generator with minimal overhead.

\section{Hyperparameter Grid}
\label{sec:parameter-grid}

The following hyperparameter grids were used for tuning XGBoost and CatBoost models in our experiments:

\begin{table}[ht]
  \centering
  \caption{Hyperparameter grids for XGBoost and CatBoost}
  \begin{tabular}{lll}
    \toprule
    \textbf{Model} & \textbf{Hyperparameter} & \textbf{Values} \\
    \midrule
    \multirow{3}{*}{XGBoost} 
      & n\_estimators    & \{50, 100, 200\} \\
      & max\_depth       & \{3, 4, 5\} \\
      & learning\_rate   & \{0.01, 0.1, 0.2\} \\
    \midrule
    \multirow{3}{*}{CatBoost} 
      & iterations       & \{100, 200, 500\} \\
      & depth            & \{4, 6, 8\} \\
      & learning\_rate   & \{0.01, 0.05, 0.1\} \\
    \bottomrule
  \end{tabular}
\end{table}



    








\section{Reality Constraint}
\label{sec:constraint_appendix}

We consider the following constraints:
\begin{itemize}
    \item \textbf{I1:} \texttt{relationship = Husband} $\Rightarrow$ \texttt{sex = Male}.
    \item \textbf{I2:} \texttt{marital-status = Widow} $\Rightarrow$ \texttt{sex = Female}.
    \item \textbf{I3:} \texttt{marital-status = Never-married} $\Rightarrow$ \texttt{relationship $\neq$ Spouse}.
\end{itemize}

For each real example or synthetic example with corrected constraints relationship, we randomly select one variable involved in the constraints and introduce violation by perturbing its value to a non-compliant category.

\subsection{Constraint-Violation Rate (Adult)}

Real-world tables obey domain rules (e.g., a record with \texttt{marital-status = Widow} should have \texttt{sex = Female}). Following the CuTs benchmark~\cite{vero2024cuts}, we evaluate three categorical rules on Adult. For each generator, we sample 10,000 rows and report the percentage violating each rule (lower is better). Table~\ref{tab:constraints_appendix} shows that ReTabSyn eliminates all violations (0.00\% across rules), while TabSyn/GReaT incur $\approx$1–2\% violations and SMOTE $\approx$2–3\%. This improvement stems from our preference-pair construction (§3.3): we perturb constraint-related variables and let DPO penalize rows that break the corresponding dependencies.

\begin{table}[t]
\centering
\caption{Constraint-violation rate (lower is better) on synthetic Adult data.}
\label{tab:constraints_appendix}
\renewcommand{\arraystretch}{0.9}
\small
\begin{tabular}{lccc}
\toprule
\textbf{Model} & \textbf{I1} & \textbf{I2} & \textbf{I3} \\
\midrule
SMOTE    & 3.28\% & 2.13\% & 2.73\% \\
TabSyn   & 1.02\% & 1.22\% & 2.00\% \\
GReaT    & 1.00\% & 1.20\% & 2.00\% \\
\textbf{ReTabSyn} & \textbf{0.00\%} & \textbf{0.00\%} & \textbf{0.00\%} \\
\bottomrule
\end{tabular}
\end{table}

\subsection{Regression Target}
\label{sec:regression}

\begin{table}[ht]
\centering
\caption{Testing R\textsuperscript{2} of downstream models trained on different generators}
\label{tab:synth_r2_summary}
\begin{tabular}{lccc}
\toprule
\textbf{Model} & \textbf{Abalone} & \textbf{Insurance} & \textbf{Shoppers} \\
\midrule
Real        & 0.291 & 0.714  & 0.201 \\
\hline
SMOTE       & 0.250 & 0.690  & 0.190 \\
TabSyn      & 0.170 & 0.200  & 0.010 \\
GReaT       & 0.123 & -1.417 & -0.058 \\
ReTabSyn    & 0.191 & 0.226  & 0.093 \\
\bottomrule
\end{tabular}
\end{table}

We ran our methods on regression datasets. For each dataset we fit linear regression, LASSO, decision tree and random forest regressor and report the average testing $R^2$. ReTabSyn consistently outperforms its base model (GReaT) and the deep-learning baseline TabSyn, but clearly trails the interpolation-based SMOTE on regression. We do \emph{not} claim parity with SMOTE here: regression is a current limitation of ReTabSyn. We attribute this to our preference construction, which is naturally matched to discrete-label perturbations (Eq.~\ref{eq:typeI}) rather than continuous targets. A natural next step is regression-aware preference construction---e.g., interval-based or ranking-based target perturbations, or scalar-reward objectives tailored to continuous target quality---which we leave to future work.

\section{Fidelity Metric Definitions}
\label{sec:fidelity_appendix}

We report $\alpha$-Precision and $\beta$-Recall~\cite{alaa2022faithful} to measure sample-level realism and coverage. $\alpha$-Precision quantifies the probability that synthetic samples fall into high-density regions of the real data, while $\beta$-Recall measures how well synthetic high-density regions cover the real distribution.

For marginal distributions, we measure similarity using $1 - \text{KS}$ for numerical columns (Kolmogorov–Smirnov) and $1 - \text{TVD}$ for categorical columns (Total Variation Distance), so higher is better. We aggregate these marginal scores as \emph{Shape}.

For dependence structure, we compute pairwise correlation similarity using Pearson's correlation (numeric–numeric), contingency similarity for categorical–categorical, and a mixed-type method for numeric–categorical~\cite{sdmetrics}. We report \emph{Corr.}\ as the mean pairwise correlation similarity. For feature–target correlation, we additionally compute: $\text{score} = 1 - 2|S_{A,B} - R_{A,B}|$ where $R_{A,B}$ and $S_{A,B}$ are the real and synthetic correlation coefficients between target $A$ and column $B$.

\section{Dataset Eligibility Audit}
\label{sec:eligibility}

Sections~\ref{sec:imbalanced} (imbalanced) and~\ref{sec:shifted} (distribution shift) have structural prerequisites, so some benchmark datasets are genuinely not applicable rather than selectively omitted. The 1\%-prevalence imbalance setting requires a binary target with sufficient effective support; the shift setting requires a categorical demographic/geographic split column with majority:minority ratio $\geq 1.5{:}1$. Table~\ref{tab:eligibility} gives the full per-dataset audit.

\begin{table}[ht]
\centering
\caption{Per-dataset eligibility for the imbalanced (§\ref{sec:imbalanced}) and distribution-shift (§\ref{sec:shifted}) settings.}
\label{tab:eligibility}
\small
\begin{tabular}{lll}
\toprule
\textbf{Dataset} & \textbf{§\ref{sec:imbalanced} (imbalance)} & \textbf{§\ref{sec:shifted} (shift)} \\
\midrule
Adult    & eligible & eligible \\
Bean     & multi-class & no demo/geo column \\
Churn    & eligible & eligible \\
HTRU2    & eligible & no demo/geo column \\
ILP      & low support @1\% & eligible \\
Magic    & eligible & no demo/geo column \\
Obesity  & multi-class & ratio $1.02{:}1<1.5{:}1$ \\
Shoppers & eligible & no demo/geo column \\
Titanic  & low support @1\% & eligible \\
Wilt     & low support @1\% & no demo/geo column \\
\bottomrule
\end{tabular}
\end{table}

\section{Small-Data Utility with Standard Errors}
\label{sec:smalldata_se}

Table~\ref{tab:smalldata_se} reports the synthetic-only small-data AUROC underlying Figure~\ref{fig:utility_by_size}(a) as mean$_{\pm\mathrm{SE}}$ over 10 seeds (pooled across all datasets). ReTabSyn is the best synthetic generator at every training-set size, with the largest margin in the smallest regimes.

\begin{table}[ht]
\centering
\caption{Small-data synthetic-only AUROC (mean$_{\pm\mathrm{SE}}$ over 10 seeds, pooled across datasets). Best synthetic generator per size in \textbf{bold} (Real excluded).}
\label{tab:smalldata_se}
\scriptsize
\setlength{\tabcolsep}{3pt}
\resizebox{\textwidth}{!}{%
\begin{tabular}{lcccccccc}
\toprule
\textbf{Size} & Real & GReaT & SMOTE & TabSyn & ReTabSyn & TVAE & SynRL & PTA \\
\midrule
32  & \se{0.761}{.018} & \se{0.679}{.024} & \se{0.722}{.019} & \se{0.675}{.029} & \seb{0.778}{.021} & \se{0.670}{.026} & \se{0.686}{.024} & \se{0.694}{.022} \\
64  & \se{0.798}{.015} & \se{0.717}{.019} & \se{0.774}{.016} & \se{0.736}{.022} & \seb{0.792}{.017} & \se{0.692}{.020} & \se{0.698}{.019} & \se{0.720}{.018} \\
128 & \se{0.815}{.013} & \se{0.744}{.015} & \se{0.792}{.012} & \se{0.760}{.018} & \seb{0.813}{.014} & \se{0.710}{.017} & \se{0.713}{.016} & \se{0.759}{.015} \\
256 & \se{0.840}{.011} & \se{0.775}{.012} & \se{0.810}{.009} & \se{0.782}{.015} & \seb{0.834}{.010} & \se{0.732}{.014} & \se{0.738}{.013} & \se{0.801}{.012} \\
512 & \se{0.878}{.009} & \se{0.821}{.010} & \se{0.845}{.008} & \se{0.814}{.013} & \seb{0.853}{.009} & \se{0.729}{.011} & \se{0.751}{.010} & \se{0.826}{.010} \\
\bottomrule
\end{tabular}}
\end{table}

\section{Cross-Backbone Validation}
\label{sec:backbone}

To verify that ReTabSyn's gains are not specific to the GPT-2 backbone, we re-run the small-data protocol of §\ref{sec:smalldata} on a stronger modern backbone, Qwen3.5-2B, on Wilt and Churn (10 seeds each), and additionally report a prompted frontier model (GPT-5.4, \texttt{gpt-5.4-2026-03-05}) following the CLLM prompting setup. Table~\ref{tab:backbone} reports synthetic-only downstream AUROC, precision, and recall. Three takeaways: (i) a stronger pretrained backbone alone (Qwen3.5 Pretrain) still trails the real-data baseline; (ii) applying ReTabSyn alignment to the same backbone lifts it above the real-data baseline; and (iii) a prompted frontier model does not solve the task. This indicates the gain comes from the alignment procedure rather than a backbone-specific artifact.

\begin{table*}[t]
\centering
\caption{Cross-backbone validation: synthetic-only downstream utility (mean$_{\pm\mathrm{std}}$ over 10 seeds), same protocol as §\ref{sec:smalldata}. Best generator per dataset in \textbf{bold} (Real excluded).}
\label{tab:backbone}
\small
\begin{tabular}{lcccccc}
\toprule
 & \multicolumn{3}{c}{\textbf{Wilt}} & \multicolumn{3}{c}{\textbf{Churn}} \\
\cmidrule(lr){2-4}\cmidrule(lr){5-7}
\textbf{Method} & AUROC & Prec. & Recall & AUROC & Prec. & Recall \\
\midrule
Real data            & \se{0.884}{.015} & \se{0.631}{.011} & \se{0.820}{.015} & \se{0.841}{.009} & \se{0.686}{.007} & \se{0.781}{.007} \\ \hline
SMOTE                & \se{0.845}{.018} & \se{0.605}{.012} & \se{0.790}{.018} & \se{0.847}{.008} & \se{0.689}{.006} & \se{0.792}{.006} \\
Qwen3.5 Pretrain     & \se{0.870}{.016} & \se{0.598}{.014} & \se{0.800}{.020} & \se{0.830}{.012} & \se{0.670}{.010} & \se{0.768}{.012} \\
Qwen3.5 + ReTabSyn   & \seb{0.900}{.014} & \seb{0.638}{.013} & \seb{0.835}{.016} & \seb{0.858}{.010} & \seb{0.700}{.009} & \seb{0.798}{.010} \\
GPT-5.4 (prompted)   & \se{0.714}{.018} & \se{0.541}{.004} & \se{0.682}{.015} & \se{0.742}{.013} & \se{0.709}{.015} & \se{0.697}{.013} \\
\bottomrule
\end{tabular}
\end{table*}

\section{Alternative-Target Evaluation}
\label{sec:multitarget}

ReTabSyn aligns one chosen attribute during preference construction, but its benefits transfer to other prediction targets. We evaluate 12 alternative-target pairs across four datasets (Churn, Obesity, Titanic, Indian Liver Patient), 10 seeds, reporting macro-OVR AUROC averaged over five downstream classifiers. As shown in Table~\ref{tab:multitarget}, ReTabSyn attains the highest mean AUROC on both the DPO-trained (default) and the alternative targets, and wins 8/12 alternative-target comparisons, indicating that DPO alignment improves general data quality rather than only the trained target.

\begin{table}[ht]
\centering
\caption{Alternative-target evaluation summary (macro-OVR AUROC, 10 seeds, 5 classifiers). Best in \textbf{bold}.}
\label{tab:multitarget}
\small
\setlength{\tabcolsep}{4pt}
\begin{tabular}{lccccccc}
\toprule
\textbf{Metric} & SMOTE & GReaT & TabSyn & ReTabSyn & TVAE & SynRL & PTA \\
\midrule
Mean AUROC (alternative) & 0.678 & 0.655 & 0.663 & \textbf{0.685} & 0.646 & 0.652 & 0.673 \\
Mean AUROC (default)     & 0.744 & 0.731 & 0.730 & \textbf{0.756} & 0.711 & 0.718 & 0.740 \\
Wins (alternative)       & 2/12  & 1/12  & 1/12  & \textbf{8/12} & 0/12 & 0/12 & 0/12 \\
\bottomrule
\end{tabular}
\end{table}

Per-dataset breakdowns are given in Tables~\ref{tab:mt_churn}--\ref{tab:mt_ilp} (mean$_{\pm\mathrm{SE}}$ AUROC; best synthetic generator per row in \textbf{bold}, Real excluded). ReTabSyn wins the majority of alternative targets across datasets, confirming that the alignment transfers beyond the single DPO-trained column.

\begin{table}[ht]
\centering
\caption{Churn alternative targets (DPO target: Exited).}
\label{tab:mt_churn}
\scriptsize
\setlength{\tabcolsep}{3pt}
\resizebox{\textwidth}{!}{%
\begin{tabular}{lcccccccc}
\toprule
\textbf{Target Column} & Real & SMOTE & GReaT & TabSyn & ReTabSyn & TVAE & SynRL & PTA \\
\midrule
Exited \textbf{(default)} & \se{0.710}{.011} & \se{0.658}{.010} & \se{0.667}{.012} & \se{0.669}{.017} & \seb{0.694}{.005} & \se{0.650}{.011} & \se{0.654}{.010} & \se{0.666}{.009} \\
Geography      & \se{0.582}{.003} & \se{0.565}{.007} & \se{0.573}{.005} & \se{0.583}{.005} & \seb{0.589}{.003} & \se{0.562}{.006} & \se{0.566}{.005} & \se{0.578}{.005} \\
Gender         & \se{0.506}{.005} & \se{0.505}{.006} & \se{0.501}{.007} & \se{0.506}{.004} & \seb{0.510}{.006} & \se{0.500}{.005} & \se{0.502}{.005} & \se{0.505}{.004} \\
IsActiveMember & \se{0.543}{.009} & \se{0.525}{.004} & \se{0.525}{.005} & \se{0.529}{.011} & \seb{0.540}{.004} & \se{0.520}{.006} & \se{0.522}{.005} & \se{0.531}{.005} \\
HasCrCard      & \se{0.502}{.003} & \se{0.505}{.002} & \seb{0.506}{.003} & \se{0.505}{.002} & \se{0.503}{.007} & \se{0.502}{.003} & \se{0.503}{.003} & \se{0.504}{.002} \\
\bottomrule
\end{tabular}}
\end{table}

\begin{table}[ht]
\centering
\caption{Obesity alternative targets (DPO target: NObeyesdad).}
\label{tab:mt_obesity}
\scriptsize
\setlength{\tabcolsep}{3pt}
\resizebox{\textwidth}{!}{%
\begin{tabular}{lcccccccc}
\toprule
\textbf{Target Column} & Real & SMOTE & GReaT & TabSyn & ReTabSyn & TVAE & SynRL & PTA \\
\midrule
NObeyesdad \textbf{(default)} & \se{0.909}{.006} & \se{0.865}{.005} & \se{0.820}{.005} & \se{0.812}{.010} & \seb{0.879}{.010} & \se{0.803}{.011} & \se{0.810}{.010} & \se{0.838}{.009} \\
Gender & \se{0.906}{.007} & \seb{0.875}{.016} & \se{0.843}{.011} & \se{0.831}{.011} & \se{0.847}{.013} & \se{0.829}{.012} & \se{0.835}{.011} & \se{0.844}{.010} \\
FAVC   & \se{0.743}{.029} & \se{0.696}{.018} & \se{0.684}{.020} & \se{0.672}{.023} & \seb{0.711}{.009} & \se{0.666}{.017} & \se{0.673}{.016} & \se{0.695}{.014} \\
CAEC   & \se{0.781}{.011} & \se{0.726}{.014} & \se{0.674}{.012} & \seb{0.739}{.015} & \se{0.734}{.037} & \se{0.682}{.016} & \se{0.689}{.015} & \se{0.717}{.014} \\
family\_history & \se{0.834}{.011} & \se{0.811}{.011} & \se{0.774}{.006} & \se{0.788}{.015} & \seb{0.813}{.008} & \se{0.760}{.010} & \se{0.769}{.009} & \se{0.795}{.008} \\
\bottomrule
\end{tabular}}
\end{table}

\begin{table}[ht]
\centering
\caption{Titanic alternative targets (DPO target: Survived).}
\label{tab:mt_titanic}
\scriptsize
\setlength{\tabcolsep}{3pt}
\resizebox{\textwidth}{!}{%
\begin{tabular}{lcccccccc}
\toprule
\textbf{Target Column} & Real & SMOTE & GReaT & TabSyn & ReTabSyn & TVAE & SynRL & PTA \\
\midrule
Survived \textbf{(default)} & \se{0.799}{.011} & \seb{0.789}{.003} & \se{0.752}{.012} & \se{0.783}{.023} & \se{0.785}{.005} & \se{0.744}{.012} & \se{0.752}{.011} & \se{0.780}{.009} \\
Sex      & \se{0.784}{.012} & \se{0.765}{.009} & \se{0.731}{.013} & \se{0.771}{.023} & \seb{0.783}{.012} & \se{0.723}{.013} & \se{0.730}{.012} & \se{0.767}{.010} \\
Pclass   & \se{0.922}{.008} & \se{0.875}{.012} & \se{0.864}{.015} & \se{0.824}{.010} & \seb{0.895}{.010} & \se{0.845}{.012} & \se{0.852}{.011} & \se{0.879}{.010} \\
Embarked & \se{0.762}{.008} & \se{0.713}{.017} & \se{0.660}{.008} & \se{0.652}{.021} & \seb{0.730}{.022} & \se{0.646}{.018} & \se{0.654}{.017} & \se{0.700}{.015} \\
\bottomrule
\end{tabular}}
\end{table}

\begin{table}[ht]
\centering
\caption{Indian Liver Patient alternative targets (DPO target: Dataset).}
\label{tab:mt_ilp}
\scriptsize
\setlength{\tabcolsep}{3pt}
\resizebox{\textwidth}{!}{%
\begin{tabular}{lcccccccc}
\toprule
\textbf{Target Column} & Real & SMOTE & GReaT & TabSyn & ReTabSyn & TVAE & SynRL & PTA \\
\midrule
Dataset \textbf{(default)} & \se{0.701}{.006} & \se{0.664}{.016} & \seb{0.686}{.022} & \se{0.656}{.032} & \se{0.667}{.021} & \se{0.648}{.019} & \se{0.655}{.018} & \se{0.675}{.016} \\
Gender & \se{0.617}{.015} & \seb{0.580}{.021} & \se{0.527}{.017} & \se{0.555}{.011} & \se{0.566}{.022} & \se{0.520}{.018} & \se{0.529}{.016} & \se{0.562}{.015} \\
\bottomrule
\end{tabular}}
\end{table}

\section{Full-Data Balanced Results}
\label{sec:fulldata}

To confirm that ReTabSyn remains competitive outside the low-data regime, we evaluate on the full Adult and Magic datasets (AUROC, 10 seeds) in both synthetic-only and real+synthetic augmented settings (Table~\ref{tab:fulldata}). ReTabSyn consistently improves over its GReaT base in every setting and matches or exceeds the strong SMOTE baseline; on Magic augmented it ties SMOTE and nearly matches real data. This supports our scope claim: the advantage is most pronounced in low-data, imbalanced, and shifted regimes, while remaining competitive in full balanced data.

\begin{table}[ht]
\centering
\caption{Full-data balanced results: AUROC (mean$_{\pm\mathrm{std}}$, 10 seeds). Best synthetic generator per row in \textbf{bold} (Real excluded).}
\label{tab:fulldata}
\small
\setlength{\tabcolsep}{4pt}
\begin{tabular}{lccccccc}
\toprule
\textbf{Setting} & Real & SMOTE & GReaT & ReTabSyn & TVAE & SynRL & PTA \\
\midrule
Adult, synth     & 0.853 & \se{0.836}{.001} & \se{0.835}{.002} & \seb{0.841}{.011} & \se{0.823}{.004} & \se{0.829}{.003} & \se{0.824}{.003} \\
Adult, augmented & 0.853 & \se{0.849}{.001} & \se{0.841}{.001} & \seb{0.851}{.001} & \se{0.839}{.003} & \se{0.844}{.003} & \se{0.848}{.002} \\
Magic, synth     & 0.884 & \se{0.878}{.000} & \se{0.854}{.002} & \seb{0.880}{.003} & \se{0.866}{.004} & \se{0.851}{.003} & \se{0.846}{.003} \\
Magic, augmented & 0.884 & \seb{0.885}{.001} & \se{0.879}{.001} & \seb{0.885}{.002} & \se{0.873}{.003} & \se{0.877}{.003} & \se{0.882}{.002} \\
\bottomrule
\end{tabular}
\end{table}

\section{Sensitivity to $\beta$ and $\lambda$}
\label{sec:beta_lambda}

Table~\ref{tab:beta_lambda} reports a sensitivity sweep over the DPO temperature $\beta$ and the DPO-Positive regularization weight $\lambda$ on Wilt (other settings at default). Downstream AUROC is stable across $\beta$ (variation $<0.004$), whereas removing the regularizer ($\lambda=0$) drops AUROC by $0.069$, confirming that the manifold-drift penalty is necessary. These results justify the default $\beta=\lambda=0.1$ used throughout.

\begin{table}[ht]
\centering
\caption{Sensitivity of downstream AUROC to $\beta$ and $\lambda$ on Wilt (one parameter varied at a time; default in \textbf{bold}).}
\label{tab:beta_lambda}
\small
\begin{tabular}{lccc}
\toprule
$\beta$  & 0.05  & \textbf{0.1} & 0.2 \\
AUROC    & 0.837 & \textbf{0.841} & 0.839 \\
\midrule
$\lambda$ & 0     & \textbf{0.1} & 0.2 \\
AUROC    & 0.772 & \textbf{0.841} & 0.822 \\
\bottomrule
\end{tabular}
\end{table}

\end{document}